\documentclass[10pt,twocolumn,letterpaper]{article}

\usepackage{iccv}
\usepackage{times}
\usepackage{epsfig}
\usepackage{graphicx}
\usepackage{amsmath}
\usepackage{amssymb}
\usepackage{booktabs}
\usepackage{mathabx}

\usepackage{comment} 
\usepackage{boldline} 
\usepackage[]{caption}
\usepackage{soul}

\usepackage{multirow}

\usepackage[pagebackref=true,breaklinks=true,letterpaper=true,colorlinks,bookmarks=false]{hyperref}

\iccvfinalcopy 

\def\iccvPaperID{****} 

\ificcvfinal\fi

\newcommand\blfootnote[1]{%
  \begingroup
  \renewcommand\thefootnote{}\footnote{#1}%
  \addtocounter{footnote}{-1}%
  \endgroup
}

\begin{document}

\title{Is it Time to Replace CNNs with Transformers for Medical Images?}

\author{
Christos Matsoukas \textsuperscript{
$1,2,3$
\thanks{Corresponding author: Christos Matsoukas \textless{}matsou@kth.se\textgreater{}}}
\qquad
Johan Fredin Haslum \textsuperscript{$1,2,3$}
\qquad
Magnus Söderberg \textsuperscript{$3$}
\qquad
Kevin Smith \textsuperscript{$1,2$}
\\\\
\textsuperscript{$1$} KTH Royal Institute of Technology, Stockholm, Sweden \\
\textsuperscript{$2$} Science for Life Laboratory, Stockholm, Sweden \\
\textsuperscript{$3$} AstraZeneca, Gothenburg, Sweden \\
}

\maketitle
\ificcvfinal\thispagestyle{empty}\fi

\begin{abstract}

Convolutional Neural Networks (CNNs) have reigned for a decade as the de facto approach to automated medical image diagnosis. 
Recently, vision transformers (ViTs) have appeared as a competitive alternative to CNNs, yielding similar levels of performance while possessing several interesting properties that could prove beneficial for medical imaging tasks.
In this work, we explore whether it is time to move to transformer-based models or if we should keep working with CNNs -- can we trivially switch to transformers? 
If so, what are the advantages and drawbacks of switching to ViTs for medical image diagnosis?
We consider these questions in a series of experiments on three mainstream medical image datasets.
Our findings show that, while CNNs perform better when trained from scratch, off-the-shelf vision transformers using default hyperparameters are on par with CNNs when pretrained on ImageNet, and outperform their CNN counterparts when pretrained using self-supervision.

\end{abstract}

\section{Introduction}

\blfootnote{\textit{\\Originally published at the ICCV 2021 Workshop on Computer Vision for Automated Medical Diagnosis (CVAMD).}}

Vision transformers have gained increased popularity for image recognition tasks recently, signalling a transition from convolution-based feature extractors (CNNs) to attention-based models (ViTs).
Following the success of Dosovitskiy et al.~\cite{dosovitskiy2020image}, numerous approaches for adapting transformers to vision tasks have been suggested \cite{khan2021transformers}.
In the natural image domain, transformers have been shown to outperform CNNs on standard vision tasks such as \textsc{ImageNet}  classification, \cite{dosovitskiy2020image} as well as in object detection \cite{carion2020end} and semantic segmentation \cite{ranftl2021vision}.
The attention mechanism central to transformers offers several key advantages over convolutions: (1) it captures long-range relationships, (2) it has the capacity for adaptive modeling via dynamically computed self-attention weights that capture relationships between tokens, (3) it provides a type of built-in saliency which gives insight as to what the model focused on~\cite{caron2021emerging}.

Yet, evidence suggests that vision transformers require very large datasets to outperform CNNs -- in \cite{dosovitskiy2020image}, the benefits of ViT only became evident when Google's private 300 million image dataset, JFT-300M, was used for pretraining.
Their reliance on data of this scale is a barrier to the widespread application of transformers.
This problem is particularly acute in the medical imaging domain, where datasets are smaller and are often accompanied by less reliable labels.

CNNs, like ViTs, suffer worse performance when data is scarce.
The standard solution is to employ transfer learning:~typically, a model is pretrained on a larger dataset such as \textsc{ImageNet} \cite{imagenet_cvpr09} and then fine-tuned for specific tasks using smaller, specialized, datasets.
CNNs pre-trained on \textsc{ImageNet} typically outperform those trained from scratch in the medical domain, both in terms of final performance and reduced training time \cite{transfusion}.

Self-supervision is a learning approach to deal with unlabeled data that has recently gained much attention.
It has been shown that self-supervised pretraining of CNNs in the target domain before fine-tuning can increase performance \cite{azizi2021big}.
Initialization from \textsc{ImageNet} helps self-supervised CNNs converge faster, and usually with better predictive performance \cite{azizi2021big}.

These techniques to deal with the lack of data in the medical image domain have proven effective for CNNs, \emph{but it remains unclear whether vision transformers benefit similarly}.
Some studies suggest that pre-training CNNs for medical image analysis using \textsc{ImageNet} does not rely on feature reuse --following conventional wisdom-- but, rather due to better initialization and weight scaling \cite{transfusion}.
This calls into question whether transformers benefit from these techniques.
If they do, there is little to prevent ViTs from becoming the dominant architecture for medical images.

In this work, we explore whether ViTs can easily replace CNNs for medical imaging tasks, and if there is an advantage of doing so.
We consider the use-case of a typical practitioner, equipped with a limited computational budget and access to conventional medical datasets, with an eye towards ``plug-and-play" solutions.
To this end, we conduct experiments on three mainstream publicly available datasets.
Through these experiments we show that:
\begin{itemize}
    \item ViTs pretrained on \textsc{ImageNet} perform comparably to CNNs when data is limited.
    \item Transfer learning favours ViTs when applying standard training protocols and settings.
    \item ViTs outperform their CNN counterparts when self-supervised pre-training is followed by supervised fine-tuning.
\end{itemize}
These findings suggest that medical image analysis can seamlessly transition from CNNs to ViTs, while at the same time gaining improved explainability properties.
To promote transparency and reproducibility, we share our open-source code, available at
\href{https://github.com/ChrisMats/medical_transformers}{https://github.com/ChrisMats/medical\_transformers}.

\section{Related Work}

The use of vision transformers in the natural imaging domain has exploded recently, with applications ranging from classification \cite{dosovitskiy2020image}, to object detection \cite{carion2020end} and segmentation \cite{ranftl2021vision}.
In medical imaging, however, the use of ViTs has been limited -- primarily focused on focused on segmentation \cite{chen2021transunet,pak2020efficient}.
Only a handful of studies have tackled other tasks such as 3-D image registration \cite{chen2021vit} and detection \cite{duong2021detection}.
Notably, none of these works consider pure, off-the-shelf, vision transformers -- all propose custom architectures combining transformer/attention modules with convolutional feature extractors.

Although it is well-known that CNNs usually benefit from transfer learning to medical imaging domains, the source of these benefits is disputed.
The conventional wisdom that feature re-use contributes to better performance was questioned by Raghu et al.~\cite{transfusion}, who rather attribute the improved performance to good initialization and weight statistics.
Regardless of the reason, the question of whether ViTs benefit from transfer learning to medical domains is yet to be explored.

Recent advances in self-supervised learning have dramatically improved performance of label-free learning.
State-of-the-art methods such as DINO \cite{caron2021emerging} and BYOL \cite{grill2020bootstrap} have reached performance on par with supervised learning on \textsc{ImageNet} and other standard benchmarks.
While these top-performing methods have not yet been proven for medical imaging, Azizi et al.~\cite{azizi2021big} adopted SimCLR~\cite{chen2020simple}, an earlier self-supervised contrastive learning method, to pretrain CNNs.
This yielded state-of-the-art results for predictions on chest X-rays and skin lesions.
However, it has yet to be shown how self-supervised learning combined with ViTs performs in medical imaging, and whether this combination outperforms its CNN counterparts.

\begin{table*}[ht!]
\begin{center}
\footnotesize

\begin{tabular}{llccc}
\toprule
\textbf{Initialization} & \textbf{Model} & \textbf{APTOS2019}, $\kappa \uparrow$ & \textbf{ISIC2019}, Recall $\uparrow$ & \textbf{DDSM}, ROC-AUC $\uparrow$ 
\\
\midrule
\multirow{2}{*}{Random}
& ResNet50 & 0.849 $\pm$ 0.022 & 0.662 $\pm$ 0.018 & 0.917 $\pm$ 0.005 

\\ 
& DeiT-S   & 0.687 $\pm$ 0.017 & 0.579 $\pm$ 0.028 & 0.908 $\pm$ 0.015 

\\[0.5em] 
\multirow{2}{*}{ImageNet (supervised)} 
& ResNet50 & 0.893 $\pm$ 0.004 & 0.810 $\pm$ 0.008 & 0.953 $\pm$ 0.008 

\\ 
& DeiT-S   & 0.896 $\pm$ 0.005 & 0.844 $\pm$ 0.021 & 0.947 $\pm$ 0.011 
\\[0.5em] 
\multirow{2}{*}{\begin{tabular}[c]{@{}l@{}}ImageNet (supervised) + \\ Self-supervised with DINO \cite{caron2021emerging}\end{tabular}} 
& ResNet50 & 0.894 $\pm$ 0.008 & 0.833 $\pm$ 0.007 & 0.955 $\pm$ 0.002 
\\ 
& DeiT-S   & 0.896 $\pm$ 0.010 & 0.853 $\pm$ 0.009 & 0.956 $\pm$ 0.002 
\\ 
\bottomrule
\end{tabular}
\end{center}
\vspace{-4mm}
\caption{\emph{Comparison of vanilla CNNs vs.~ViTs with different initialization strategies on medical imaging tasks.} For each task we report the median ($\pm$ standard deviation) over 5 repetitions using the metrics that are commonly used in the literature. For \textsc{APTOS2019} we report quadratic Cohen Kappa, for \textsc{ISIC2019} we report recall
(which is semantically equivalent to the balanced multi-class accuracy)
, and for \textsc{CBIS-DDSM} we report ROC-AUC. 
}
\label{tab:finetuned}
\vspace{-3mm}
\end{table*}

\section{Methods}

The main question we investigate is whether ViTs can be used as a plug-and-play alternative to CNNs for medical diagnostic tasks.
To that end, we conducted a series of experiments to compare vanilla ViTs and CNNs under similar conditions, keeping hyperparameter tuning to a minimum.
To ensure a fair and interpretable comparison, we selected \textsc{ResNet50} \cite{resnet} as the representative CNN model, and \textsc{DeiT-S} with $16\times16$ tokens \cite{deit} as the ViT.
These models were chosen because they are comparable in the number of parameters, memory requirements, and compute.

As mentioned above, CNNs rely on initialization strategies to improve performance when data is less abundant, as is the case for medical images.
The standard approach is to use transfer learning -- initialize the model with weights pretrained on \textsc{ImageNet} and fine-tune on the target domain.
More recently, self-supervised pretraining has become a popular way to initialize neural networks \cite{grill2020bootstrap,caron2021emerging,azizi2021big}.

Accordingly, we consider three initialization strategies: \emph{(1)} randomly initialized weights, \emph{(2)} transfer learning using supervised \textsc{ImageNet} pretrained weights, \emph{(3)} self-supervised pretraining on the target dataset, after initialization as in (2).
We apply these strategies on three standard  medical imaging datasets chosen to cover a diverse set of target domains:

\vspace{-1mm}
\begin{itemize}
\vspace{-1mm}
\item \textbf{APTOS 2019} -- In this dataset, the task is classification of diabetic retinopathy images into 5 categories of disease severity \cite{kaggle}.
APTOS 2019 contains 3,662 high-resolution retinal images.
\vspace{-1mm}
\item \textbf{ISIC 2019} -- Here, the task is to classify 25,333 dermoscopic images among nine different diagnostic categories of skin lesions \cite{tschandl2018ham10000,codella2018skin,combalia2019bcn20000}.
\vspace{-1mm}
\item \textbf{CBIS-DDSM} -- 
This dataset contains 10,239 mammography images and the task is to detect the presence of masses in the mammograms.
\vspace{-1mm}
\end{itemize}
Datasets were divided  into train/test/validation splits (80/10/10), with the exception of APTOS, which was divided 70/15/15 due to its small size.
All supervised training uses an \textsc{Adam} optimizer \cite{adam}  with a base learning rate of $10^{-4}$ with a warm-up period of 1,000 iterations.
When the validation metrics saturate, the learning rate is dropped by a factor of 10 until it reaches its final value of $10^{-6}$.
All images are resized to $256\times256$ and standard augmentations  were applied\footnote{Training augmentations include: normalization; color jitter including brightness, contrast, saturation, hue; horizontal flip; vertical flip; and random resized crops.} 
We repeat each experiment five times, and select the checkpoint with highest validation score of each run.
We use the above settings unless otherwise specified.

\section{Experiments}

\paragraph{Are randomly initialized transformers useful?}
We begin by comparing \textsc{DeiT-S} against \textsc{ResNet50} with randomly initialized weights (Kaiming initialization \cite{he2015delving}).
For these experiments, the base learning rate was set to 0.0003 following a grid search.
The results in Table \ref{tab:finetuned} indicate that in this setting, CNNs outperform ViTs by a large margin across the board.
These results are in line with previous observations in the natural image domain, where ViTs trained on limited data are outperformed by similarly-sized CNNs, a trend that was attributed to ViT's lack of inductive bias \cite{dosovitskiy2020image}.
Since most medical imaging datasets are of modest size, the usefulness of randomly initialized ViTs appears to be limited.
\vspace{-2mm}

\vspace{-2mm}
\paragraph{Does pretraining transformers on \textsc{ImageNet} work in the medical image domain?}
On medical image datasets, random initialization is rarely used in practice.
The standard procedure is to train CNNs by initializing the network with \textsc{ImageNet} pretrained weights, followed by fine-tuning on data from the target domain.
Here, we investigate if this approach can be effectively applied to vanilla ViTs.
To test this, we initialize all models with weights that have been pre-trained on \textsc{ImageNet} in a fully-supervised fashion.
We then fine-tune using the procedure described above.
The results in Table \ref{tab:finetuned} show that both CNNs and ViTs benefit significantly from \textsc{ImageNet} initialization.
In fact, ViTs appear to benefit more, as they perform on par with their CNN counterparts.
This indicates that, when initialized with \textsc{ImageNet}, \emph{CNNs can be replaced with vanilla ViTs without compromising performance}  on medical imaging tasks with modest-sized training data.
\vspace{-2mm}

\paragraph{Do transformers benefit from self-supervision in the medical image domain?}
Recent self-supervised learning schemes such as DINO \cite{caron2021emerging} and BYOL \cite{grill2020bootstrap}
approach supervised learning performance.
Moreover, if they are used for pretraining in combination with supervised fine-tuning, they can achieve a new state-of-the-art \cite{caron2021emerging,grill2020bootstrap}.
While this phenomenon has been demonstrated for CNNs and ViTs in larger data regimes, it is not clear whether self-supervised pretraining of ViTs helps for medical imaging tasks, especially on modest- and low-sized data.
To test this, we adopt the self-supervised learning scheme of DINO \cite{caron2021emerging}, which can be readily applied to both CNNs and ViTs.
DINO uses self-distillation to encourage a student and teacher network to produce similar representations given differently-augmented inputs.
Our self-supervised pretraining starts with \textsc{ImageNet} initialization, then applies self-supervised learning on the target medical domain data following the default settings suggested by the authors of the original paper \cite{caron2021emerging} -- except for three small changes:
\emph{(1)} the base learning rate was set to 0.0001, \emph{(2)} the initial weight decay is set at $10^{-5}$ and increased to $10^{-4}$ using a cosine schedule, and \emph{(3)} we used an EMA of 0.99.
The same settings were used for both CNNs and ViTs; both were pre-trained for 300 epochs using a batch size of 256, followed by fine-tuning as described in Section 3.

The results reported in Table \ref{tab:finetuned} show that both ViTs and CNNs perform better with self-supervised pretraining.
ViTs appear to outperform CNNs in this setting, albeit by a small margin.
Studies on natural images suggest the gap between ViTs and CNNs will grow with more data \cite{caron2021emerging}.
\vspace{-2mm}

\section{Discussion}
Our investigation compares the performance of vanilla CNNs and ViTs on medical image tasks under three different initialization strategies.
The results of the experiments corroborate previous findings and provide new insights.

In medical images, as was previously reported in the natural image domain, we found that CNNs outperform ViTs when trained from scratch in a low data regime \cite{dosovitskiy2020image}.
This trend appears consistently across all the datasets and fits well with the ``transformers lack inductive bias'' argument. 

Surprisingly, when initialized with supervised \textsc{ImageNet} pretrained weights, the gap between CNN and ViT performance disappears on medical tasks.
The benefits of supervised \textsc{ImageNet} pretraining on CNNs is well-known, but it was unexpected that ViTs would benefit so strongly.
To the best of our knowledge, we are the first to confirm that supervised \textsc{ImageNet} pretraining is as effective for ViTs as it is for CNNs on medical imaging tasks.
This suggests that further improvements could be gained via transfer learning from other domains more closely related to the task, as is the case for CNNs ~\cite{azizpour2015factors}.

We investigated the effect of self-supervised pre-training on the medical image domain.
Our results indicate small but consistent improvements for ViTs and CNNs.
While the best overall performance is obtained using self-supervised ViTs, interestingly in this low-data regime we do not yet see the strong advantage for self-supervision favoring ViTs that was previously reported in the natural image domain with more data, \emph{e.g.} in~\cite{caron2021emerging}

Large labeled medical image datasets are rare due to the cost of expert annotation, but it may be possible to gather large amounts of unlabeled  images.
This suggests a tantalizing opportunity to apply self-supervision on large medical image datasets, where only a small fraction are labeled.

To summarize our findings, for the medical image domain:
\vspace{-2mm}
\begin{itemize}
    \item As expected, ViTs are worse than CNNs in the low data regime if one simply trains from scratch.
    \item Transfer learning bridges the performance gap between CNNs and ViTs; performance is similar.
    \item The best performance is obtained with self-supervised pre-training + fine-tuning, where ViTs enjoy a small advantage over comparable CNNs.
\end{itemize}
\vspace{-3mm}

\paragraph{Interpretability.}
It appears that ViTs can replace CNNs for medical image tasks -- are there any other reasons for choosing ViTs over CNNs?
One should consider the added benefits of visualizing transformer attention maps.
Built-in to the self-attention mechanism of transformers is an attention map that provides, \emph{for free}, new insight into how the model makes decisions.
CNNs do not naturally lend themselves well to visualizing saliency.
Popular CNN explainability methods such as class activation maps (CAM) \cite{zhou2016learning} and Grad-CAM \cite{selvaraju2017grad} provide coarse visualizations because of pooled layers.
Transformer tokens give a finer picture of attention, and the self-attention maps explicitly model interactions between every region in the image, in contrast to the limited receptive field of the CNN.
While the difference in quality of explainability has yet to be quantified,
many have noted the qualitative improvements in interpretability afforded by transformer attention \cite{caron2021emerging}.

\begin{figure}[t]
\begin{center}
\begin{tabular}{@{}c@{}c@{}c@{}}
    \includegraphics[width=0.333\columnwidth]{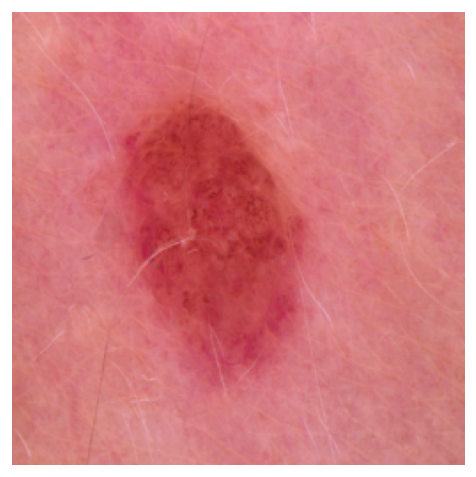} & 
    \includegraphics[width=0.333\columnwidth]{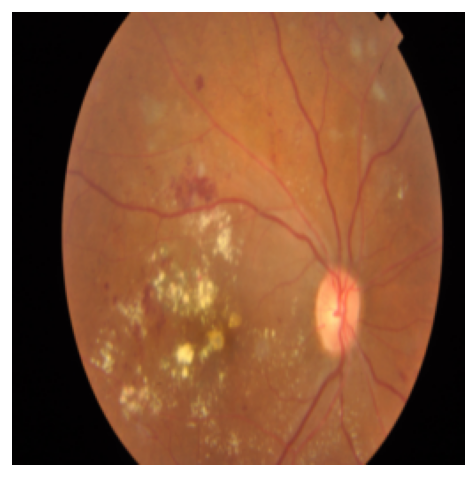} & 
    \includegraphics[width=0.333\columnwidth]{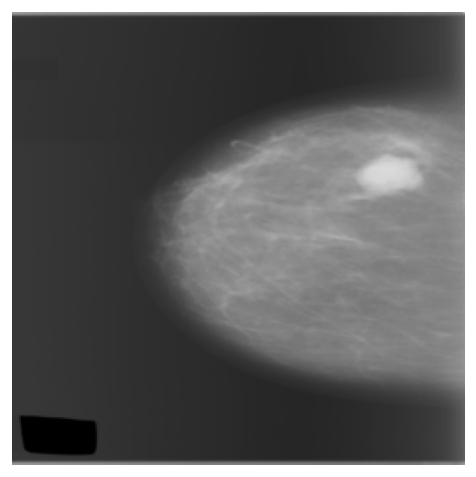}\\[-1.5mm] 
    \includegraphics[width=0.333\columnwidth]{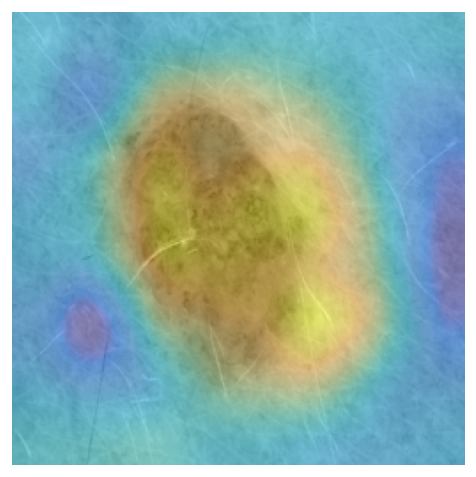} & 
    \includegraphics[width=0.333\columnwidth]{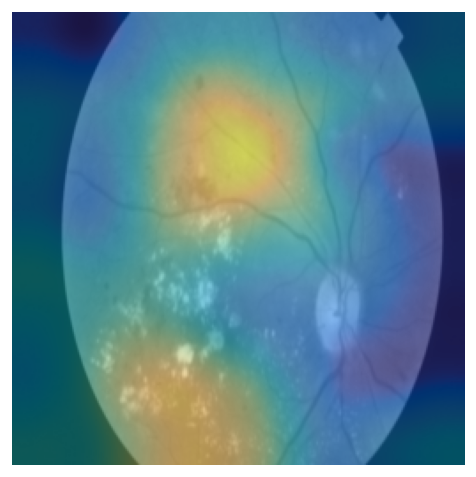} & 
    \includegraphics[width=0.333\columnwidth]{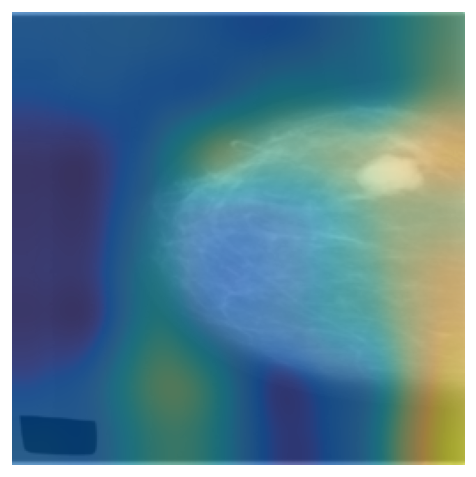}\\[-1.5mm] 
    \includegraphics[width=0.333\columnwidth]{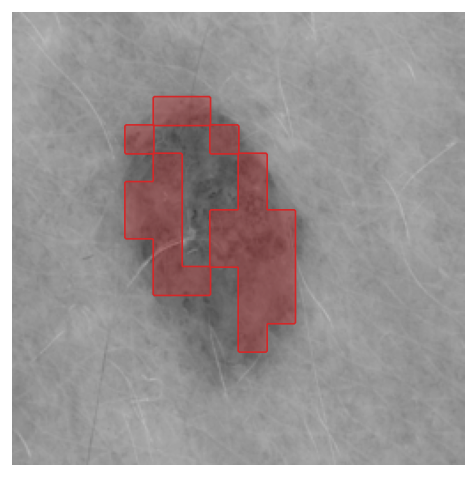} & 
    \includegraphics[width=0.333\columnwidth]{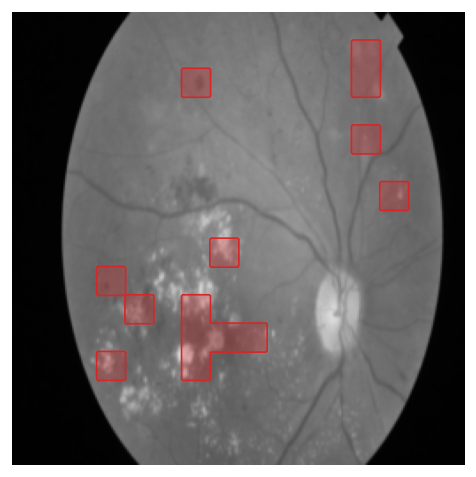} & 
    \includegraphics[width=0.333\columnwidth]{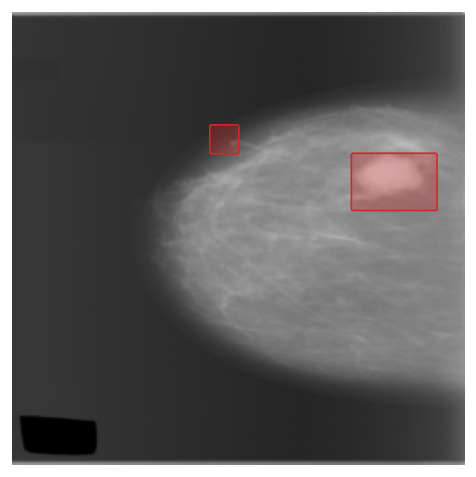}\\[-1.5mm]    
    {\scriptsize ISIC 2019} & {\scriptsize APTOS 2019} & {\scriptsize CBIS-DDSM} \\
\end{tabular}
\end{center}
\vspace{-3mm}
\caption{Comparison of saliency maps for a \textsc{ResNet50} (second row) and \textsc{DeiT-S} (third row) in three of the datasets. Each column contains the original, a visualisation of the ResNet50's Grad-CAM \cite{selvaraju2017grad} saliency and a visualisation of the \textsc{DeiT-S}'s attention map.}
\label{fig:saliency}
\vspace{-4mm}
\end{figure}

In Figure \ref{fig:saliency} we show examples from each of the datasets, along with Grad-CAM visualizations of the \textsc{ResNet-50} and the top-50\% self-attention of $16\times16$ \textsc{DeiT-S} CLS token heads.
Notice how the self-attention of ViTs provide a 
clear, localized picture of the attention, \emph{e.g.}~attention at the boundary of the skin lesion in ISIC, on hemorrhages and exudates in APTOS, and at the dense region of the breast in CBIS-DDSM.
This granularity of attention is difficult to achieve with CNNs.

\section{Conclusion}
Finally, to answer the question posed in the title: \emph{vanilla transformers can reliably replace CNNs on medical image tasks with little effort}. 
More precisely, ViTs can reach the same level of performance as CNNs in small medical datasets, but require transfer learning in order to do so. However, using ImageNet pretrained weights is the standard approach for CNNs as well, so the switch to ViTs is trivial. 
Furthermore, ViTs can outperform CNNs using SSL pre-training when working with limited number of samples, but only marginally. 
As the number of samples grows, the margin between ViT and CNN is expected to grow as well.
This equal or better performance comes with the additional benefit of built in high-resolution saliency maps that can be used to better understand the model's decisions.
\vspace{-2mm}

\paragraph{Acknowledgements.}
This work was supported by the Wallenberg Autonomous Systems Program (WASP), and
the Swedish Research Council (VR) 2017-04609.
We thank Moein Sorkhei and Emir Konuk for the thoughtful discussions.

{\small
\bibliographystyle{ieee_fullname}
\bibliography{egbib}
}

\end{document}